\documentclass[letterpaper, 10 pt, conference]{ieeeconf}  

\IEEEoverridecommandlockouts                              

\usepackage{cite}
\usepackage{amsmath,amssymb,amsfonts}
\usepackage{algorithmic}
\usepackage{graphicx}
\usepackage{textcomp}
\usepackage{xcolor}
\usepackage{float}
\usepackage{subfigure}
\usepackage{url}
\usepackage{cuted}
\usepackage[colorlinks, linkcolor=blue]{hyperref}
\usepackage{cleveref}

\title{\LARGE \bf
AcL: Action Learner for Fault-Tolerant Quadruped Locomotion Control 
}

\author{Tianyu Xu$^{1,3}$, Yaoyu Cheng$^{2,3}$, Pinxi Shen$^{2}$ and Lin Zhao$^{1,*}$
\thanks{*Corresponding author.}
\thanks{$^{1}$Tianyu Xu and Lin Zhao are with Electrical and Computer Engineering, 
        National University of Singapore, Singapore
        {\tt\small e1374408@u.nus.edu, elezhli@nus.edu.sg}}%
\thanks{$^{2}$Yaoyu Cheng and Pinxi Shen are with Mechanical Engineering, 
        National University of Singapore, Singapore
        {\tt\small, e1373292@u.nus.edu, e1373246@u.nus.edu}}%
\thanks{$^{3}$These authors contributed equally to this work.}
}

\begin{document}

\maketitle
\thispagestyle{empty}
\pagestyle{empty}


\begin{abstract}
Quadrupedal robots can learn versatile locomotion skills but remain vulnerable when one or more joints lose power. In contrast, dogs and cats can adopt limping gaits when injured, demonstrating their remarkable ability to adapt to physical conditions. Inspired by such adaptability, this paper presents Action Learner (AcL), a novel teacher-student reinforcement learning framework that enables quadrupeds to autonomously adapt their gait for stable walking under multiple joint faults. Unlike conventional teacher-student approaches that enforce strict imitation, AcL leverages teacher policies to generate style rewards, guiding the student policy without requiring precise replication. We train multiple teacher policies, each corresponding to a different fault condition, and subsequently distill them into a single student policy with an encoder-decoder architecture. While prior works primarily address single-joint faults, AcL enables quadrupeds to walk with up to four faulty joints across one or two legs, \textit{autonomously} switching between different limping gaits when faults occur. We validate AcL on a real Go2 quadruped robot under single- and double-joint faults, demonstrating fault-tolerant, stable walking, smooth gait transitions between normal and lamb gaits, and robustness against external disturbances. Videos on \url{https://github.com/ACT-legmotion/ACL-ActionLearner} and code will be made available after the paper is accepted in this repository.
\end{abstract}

\section{Introduction}
Quadruped robots are gaining popularity as versatile mobile platforms capable of navigating diverse terrains and performing robust locomotion tasks such as search and rescue operations in buildings, cargo delivery in cities, and planetary exploration. In such scenarios, quadrupeds may encounter faults that cannot be immediately repaired, requiring them to continue their tasks despite the malfunction. Consequently, quadrupeds must be capable of learning fault-tolerant gaits and adapting their locomotion in response to proprioceptive feedback when a fault occurs. However, existing robust locomotion policies often fail when internal faults disrupt the robot's performance.

This study aims to enhance fault-tolerant locomotion in quadrupedal robots, particularly when motor failures affect one or more legs. While reinforcement learning (RL) has enabled the development of robust gait control for challenging terrains, existing methods typically focus on stable locomotion in these environments without addressing the complications posed by internal motor failures \cite{rudin2022learning}\cite{cheng2024extreme}.
\begin{figure}[t]
\centerline{\includegraphics[width=1\linewidth]{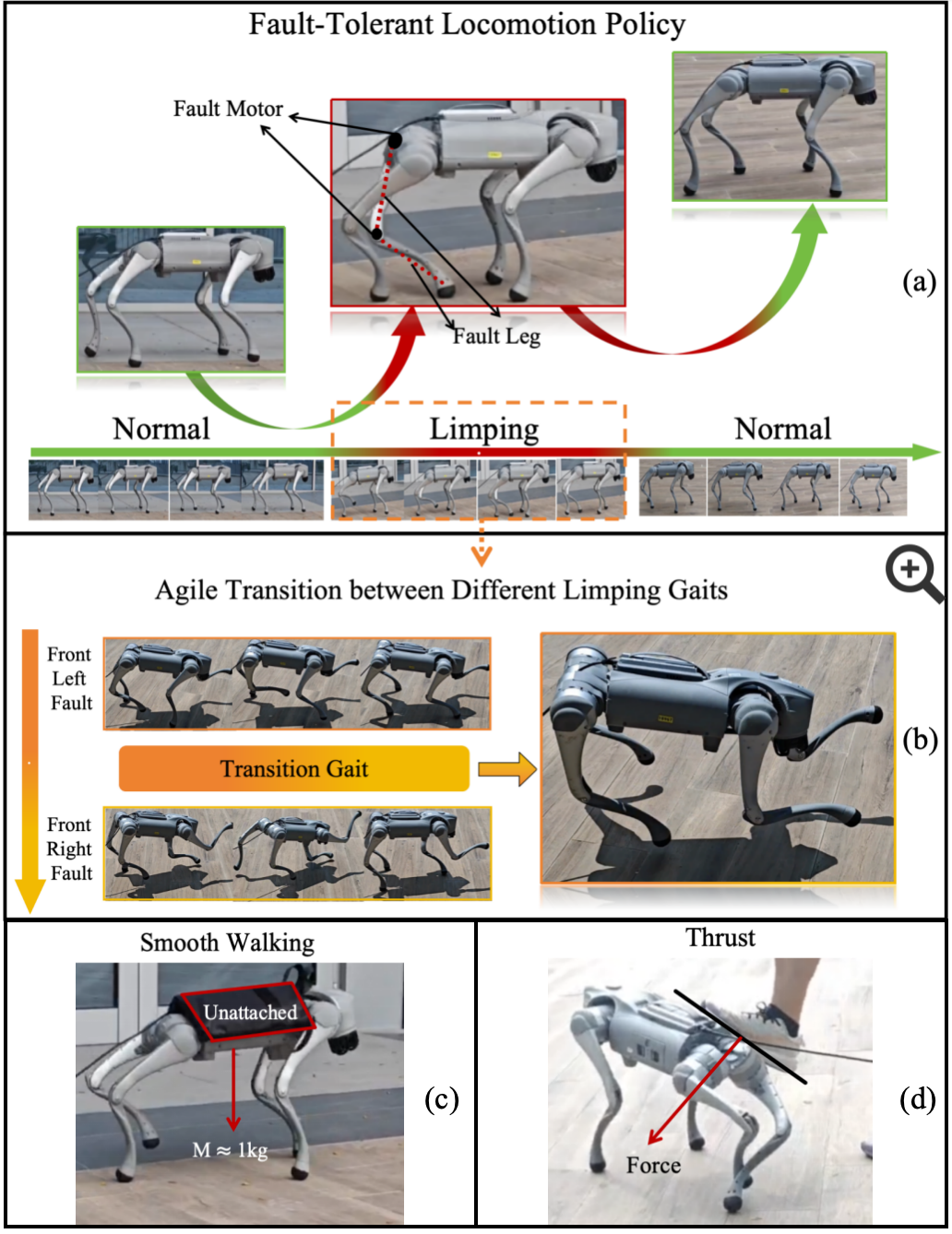}}
\caption{We propose a deep reinforcement learning controller that ensures stable and robust locomotion for a quadruped robot, even when multiple joints experience control failures. The joint faults are enabled by setting the control torque to zero (the thigh and knee joints of the left-rear leg for the figure's case). (a) The quadruped is capable of seamlessly transitioning between a normal gait and a fault-tolerant limping gait, depending on whether the affected joints are functioning or faulted. (b) Quadruped can achieve adaptively transition not only between normal and limping gaits but also among different limping gaits. (c) The limping gait demonstrates smooth walking that keeps an unattached load stable. (d) This approach also demonstrates strong adaptability and robustness even in the presence of external disturbances during the limping gait.}
\label{figtitlepic}
\end{figure}
Although some approaches have tackled fault-tolerant locomotion, they still face limitations. Most methods address only single-motor failures and lack scalability for more complex failure scenarios \cite{liu2024towards}. In contrast, our approach accommodates multiple motor failures within a single leg and ensures stable locomotion even with impairments in two legs. Moreover, current methods struggle with smooth gait transitions, often resulting in abrupt shifts between normal and fault-tolerant gaits, compromising stability. They also neglect the critical reverse transition, where recovery to a normal gait is essential \cite{anne2021meta}. Our approach addresses both transitions, ensuring seamless adaptation to faults and stable recovery.

When a quadruped experiences multiple faults, it must adapt to different gaits tailored to each specific fault scenario. Faulty joints reduce the robot's degree of freedom (DoF), altering its kinematic model and creating asymmetry. This necessitates fine-tuning each gait for natural, real-world performance. Training all fault cases within a single model may result in gaits that do not correspond to specific faults, potentially causing instability or erratic behaviors, especially when two legs fail simultaneously. Furthermore, training separate policies for each fault case and deploying them requires high-level coordination, either through an independent estimator or manual settings, limiting the generalization of the approach.

Adversarial Motion Prior (AMP) techniques \cite{peng2021amp}\cite{escontrela2022adversarial}\cite{wu2023learning} have been used to train quadrupeds with reinforcement learning, leveraging rewards based on the similarity between reference motions (such as those of a real dog) and the robot's output. Motivated by AMP, we propose a novel teacher-student RL framework, called Action Learner (AcL), to generate multiple gaits within a single policy network. In this framework, the teacher policies generate action-based rewards for the student policy, but the student is not forced to replicate the teacher's outputs or network parameters. Instead, regularization rewards are applied to enhance robustness. This approach enables the student policy to not only learn diverse gaits for various fault scenarios but also autonomously recognize faults and switch gaits accordingly, even if the teacher policies did not explicitly cover those situations.

More specifically, we propose an encoder-decoder architecture for the actor policy to facilitate fault detection and smooth gait transitions. The policy network leverages historical actions and proprioceptive observations as inputs (see~\Cref{figframework} for an illustration). Fault types are identified and encoded into a four-digit binary label, which is concatenated with the observations and fed into the decoder network to generate control actions. This design enables the policy to seamlessly switch between normal and limping gaits. We validate the trained policy on a Unitree Go2 quadruped, demonstrating stable, natural locomotion under various disturbances in real-world experiments.

In summary, the key contributions of this work are:
\begin{itemize} 
\item We propose a novel fault-tolerant policy named AcL, which enables quadrupeds to sustain locomotion with one or two faulty legs and up to four faulty joints. Real-world tests validate its ability to perform smooth gait transitions and recover to normal walking after faults are resolved. Our policy handles multiple simultaneous joint faults and adaptively transitions not only between normal and limping gaits but also among different limping gaits.
\item We develop a novel teacher-student reinforcement learning framework to distill multiple fault-tolerant gait patterns into a single policy network. The proposed framework has a great potential to develop other fast-adaptive locomotion policies across a wide range of locomotion tasks.
\item Moreover, we propose a novel encoder-decoder architecture for the actor policy which extracts the fault type information efficiently and facilitates\textit{ seamless }transitions between different gaits.

\end{itemize}

\section{Related Work}

\subsection{Reinforcement Learning in Quadruped Control}

Reinforcement learning (RL) methods have been widely applied to train robots for various tasks. Quadruped robots can be trained to perform diverse gaits, such as trotting \cite{kohl2004policy,rudin2022learning,luo2020carl}, jumping \cite{rudin2021cat,bellegarda2024robust}, and even running \cite{he2024agile}. The introduction of massively parallel reinforcement learning \cite{rudin2022learning} has significantly accelerated the training process by enabling thousands of agents to be trained simultaneously in a single iteration, enhancing the efficiency of end-to-end RL methods. Recent advancements in quadruped locomotion control have demonstrated robust policies across various terrains while requiring relatively short training times.

To mitigate the risk of damaging real robots and expedite data collection, it is common practice to first train quadrupeds in simulation and then transfer the learned policies to real platforms. In simulation, privileged observations—information unavailable in real-world scenarios—can be leveraged to facilitate training. Consequently, the teacher-student framework has been widely adopted in quadruped locomotion learning \cite{miki2022learning,jiang2023stable,cheng2024extreme,lee2020learning,vogel2024robust,yao2023learning,sorokin2022learning,ha2020quadrupedal,han2024lifelike,portelas2020teacher,shi2022reinforcement}. In this framework, the teacher policy—trained with privileged information—guides the student policy, which operates on limited observations, enabling the student policy to either learn robust behaviors or facilitate sim-to-real transfer through various distillation techniques.

\subsection{Fault-Tolerant Control for Quadruped}

Existing fault-tolerant locomotion control of quadruped robots only consider simple scenarios with a single joint fault. For example, \cite{anne2021meta} used a model-based RL algorithm to train the robot to perform a fast gait transition from normal state to the state with jammed faulty motors or a complete leg amputation. \cite{hou2024multi} demonstrates the gaits with joint power loss and joint locking via multi-task learning. \cite{liu2024towards} applied a teacher-student framework to train the quadruped robots with one faulty joint only. \cite{wu2023adaptive} designed ADAPT to train the robot to adapt to the situation where any single joint out of 12 is locked.

Our work further pushes the performance boundary to achieve robust and fast adaptive quadruped locomotion with up to four torque-losing faulty joints and can be located in either one or two legs simultaneously. In particular, the quadruped can autonomously and smoothly transition between gaits in response to different faults. We next introduce our novel method in the following sections.
\begin{figure*}[!htbp]
    \centering
    \includegraphics[width=1\linewidth]{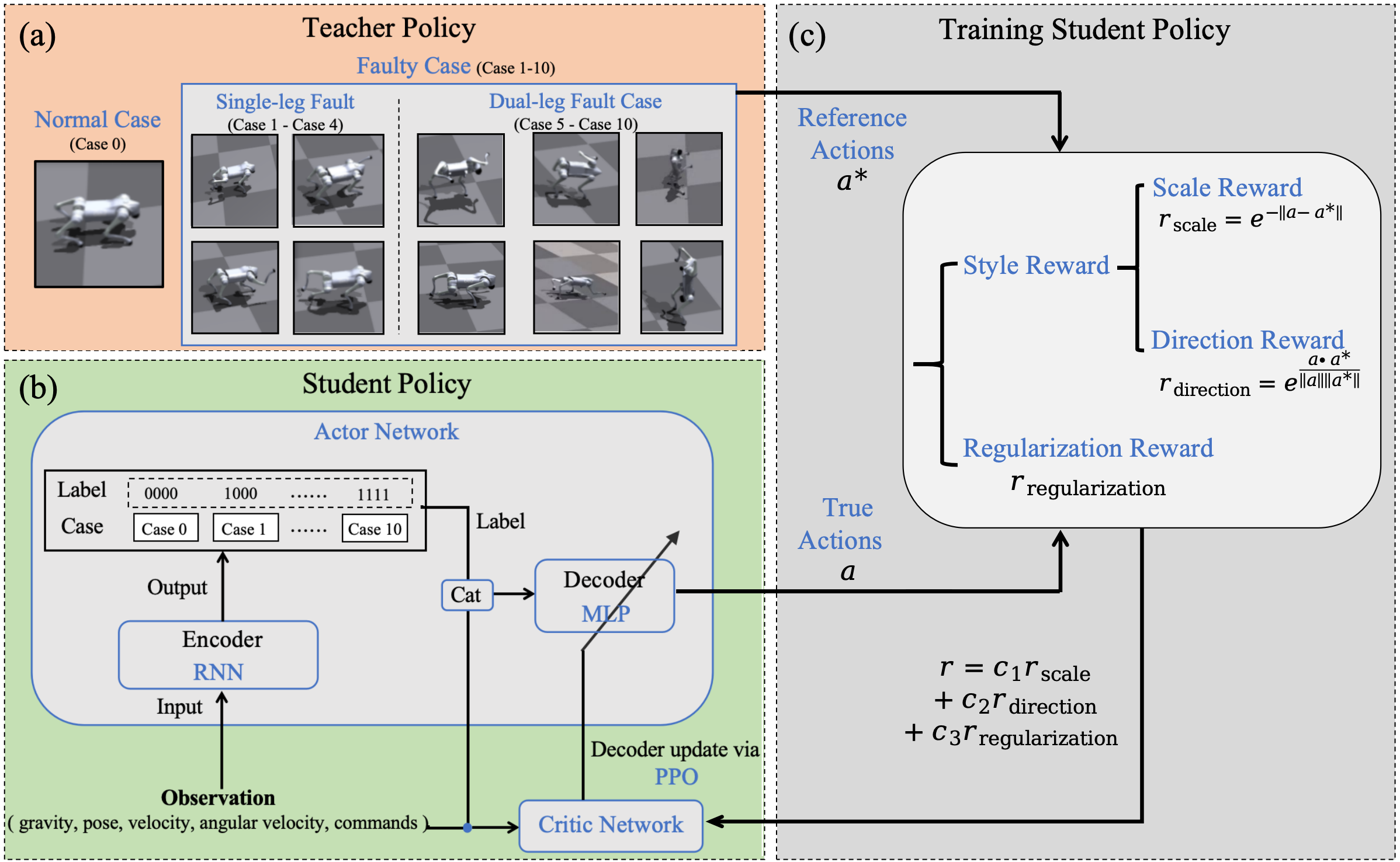}
    \caption{Overview of the proposed AcL framework — a teacher-student approach for learning multiple gaits within a single policy network. (a) The teacher policies are trained separately, each corresponding to a different fault scenario. (b) The student policy backbone uses an encoder-decoder architecture, with both the encoder and decoder trained online, but with separate parameter updates. The encoder is pre-trained using a supervised method with the datasets collected from training the teacher policies, while the decoder is pre-trained via reinforcement learning. Both encoder and decoder will be trained together online to further improve the performance. (c) The rewards consist of two components: style rewards, based on the similarity between the teacher and student policies, and regularization rewards, aimed at ensuring robust locomotion. The trained agent can autonomously and smoothly switch between different fault scenarios.}
    \label{figframework}
\end{figure*}
\section{Methods}
\subsection{AcL Overview: A Novel Teacher-Student RL Training Framework}
Training a single policy network to learn different gaits \textit{in a single stage of RL training} can be extremely challenging, if not impossible. Because each fault case requires significantly different reward scales and even distinct reward types. To tackle the challenge, we proposed a teacher-student two-stage RL training framework as illustrated in~\Cref{figframework}. We first train multiple teacher policies with each corresponding to a different fault case. The teacher policy is a simple three-layer multi-layer perceptron (MLP). It can be easily trained and fine-tuned by end-to-end RL with tailored training methods, domain randomization, and reward functions. The student policy has more complex structures. It consists of an encoder-decoder backbone, where the encoder identify and classify the fault cases based on the proprioceptive measurements, and the decoder generates the action. The encoder is trained by supervised learning, while the decoder is trained separately by RL with the guidance from the teacher policies.

Due to the structural differences between the teacher and student policy networks, direct parameter copying is infeasible. Moreover, while each teacher policy is specialized for a single fault scenario, the student policy is designed to handle more complex and diverse fault cases. As a result, the student policy does not simply mimic the actions of the teacher policies. Instead, the teacher policies provide reference actions that are used to generate style rewards, shaping a portion of the reinforcement learning objective. Meanwhile, regularization rewards are applied throughout the student policy training to encourage stability and smooth transitions, which will be introduced shortly in the sequel.
\begin{figure}[!htbp]
    \centering
    \subfigure[Normal Case: Quadruped’s diagonal leg pairs alternately contact the ground, performing a trotting gait.]{
        \includegraphics[width=0.95\linewidth]{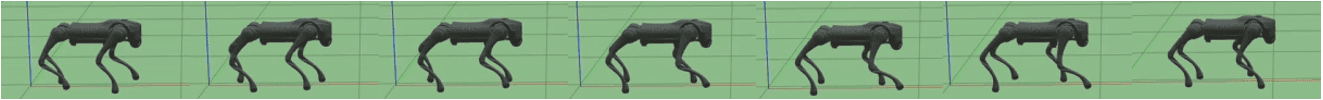}
    }
    \subfigure[Single-leg Faulty Case: Quadruped’s faulty leg is lifted, while the three normal feet form a symmetric-triangular support base to maintain stability.]{
        \includegraphics[width=0.95\linewidth]{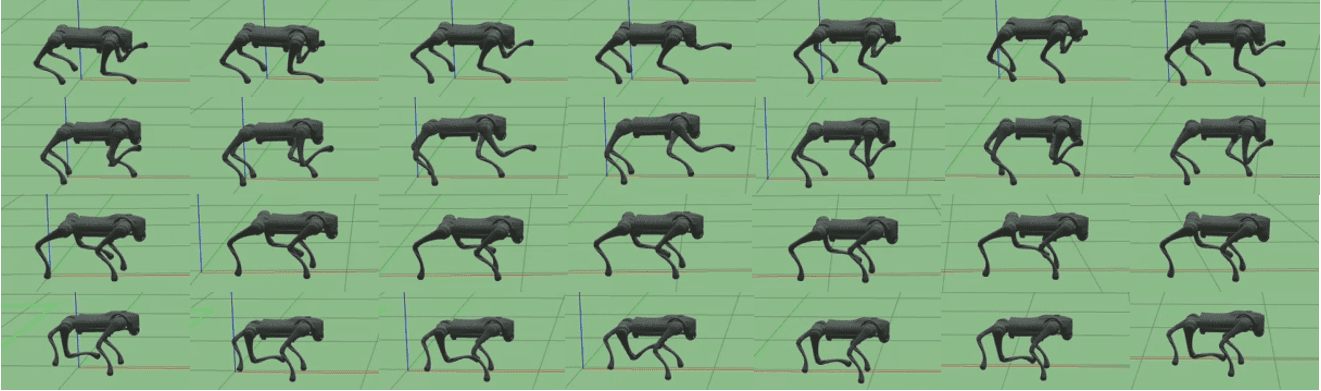}
    }
    \subfigure[Dual-leg Faulty Case: Quadruped’s faulty legs are lifted and the base is relatively straight to keep balance.]{
        \includegraphics[width=0.95\linewidth]{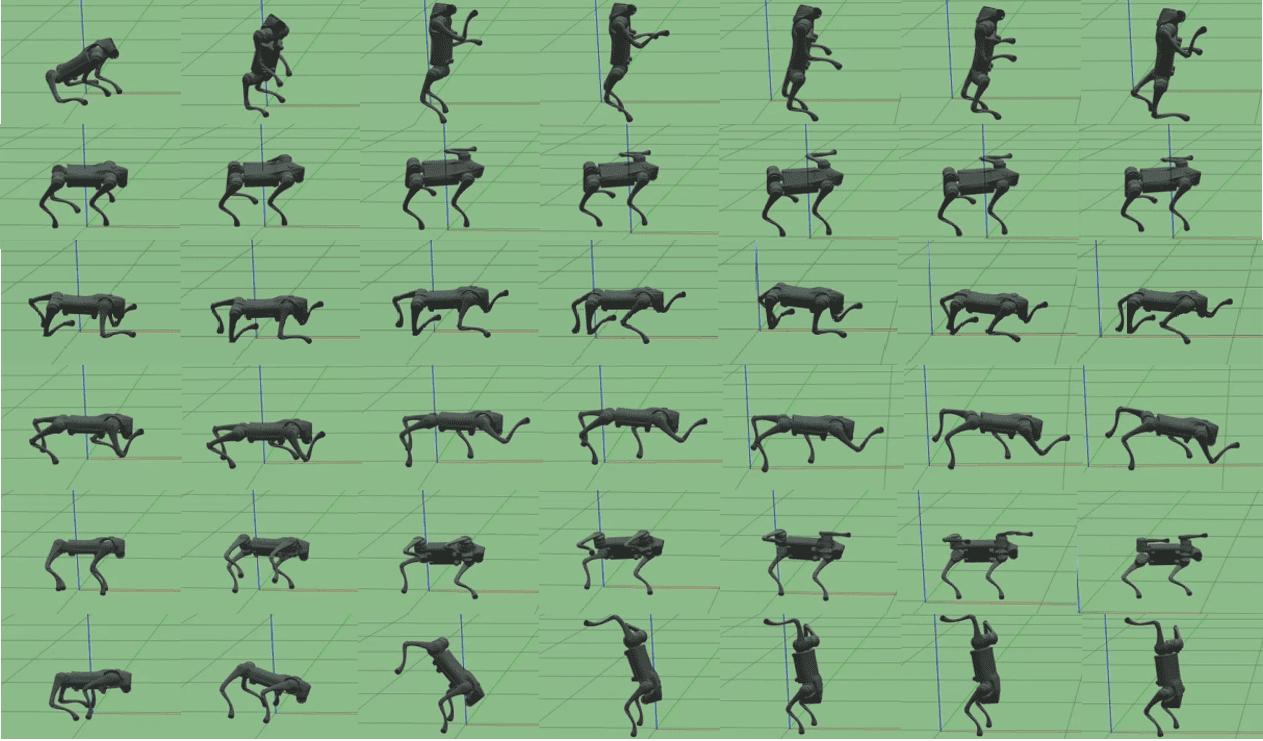}
    }
    \caption{Quadruped with teacher policies deployed in Gazebo for all 11 cases.}
    \label{fig:combinedteacherpolicies}
\end{figure}
\subsection{Student Policy Backbone}
Learning distinct gaits for various fault cases poses a significant challenge for the policy network, especially without an independent state estimator to accurately identify fault conditions. To address this limitation, we propose an encoder-decoder architecture as the student actor policy network.

Meanwhile, training a single policy network end-to-end via on-policy reinforcement learning (RL) to both identify and classify faults and generate appropriate actions for each case is inherently difficult. Therefore, we adopt a decoupled approach, where the encoder and decoder are trained separately.

The encoder is designed to extract information from historical observations, making Recurrent Neural Networks (RNNs) a natural choice. We pre-train the encoder using supervised learning to map observations to manually assigned fault case labels. The pre-training dataset is collected from the teacher policies and covers all 11 distinct fault cases.

The decoder is first pre-trained independently using an RL method, where the quadruped is subjected to various fault scenarios with periodic switches between normal and faulty states. During this phase, the ground truth fault case label (a four-digit binary code, see~\Cref{figframework}) is concatenated with the original observation input to help the decoder learn fault-specific action generation.

Moreover, to improve the performance, the pre-trained encoder and decoder are further trained together online in simulation. This joint training phase exposes the encoder to additional fault cases and enhances its accuracy in dynamic conditions. Throughout this process, the quadruped's state is periodically switched between normal and faulty modes. The encoder is refined via supervised learning using the actual fault type as ground truth, while the decoder is optimized using RL.

\begin{figure}[t]
    \centering
    \subfigure[Faults on Left Front Leg]{
        \includegraphics[width=0.45\linewidth]{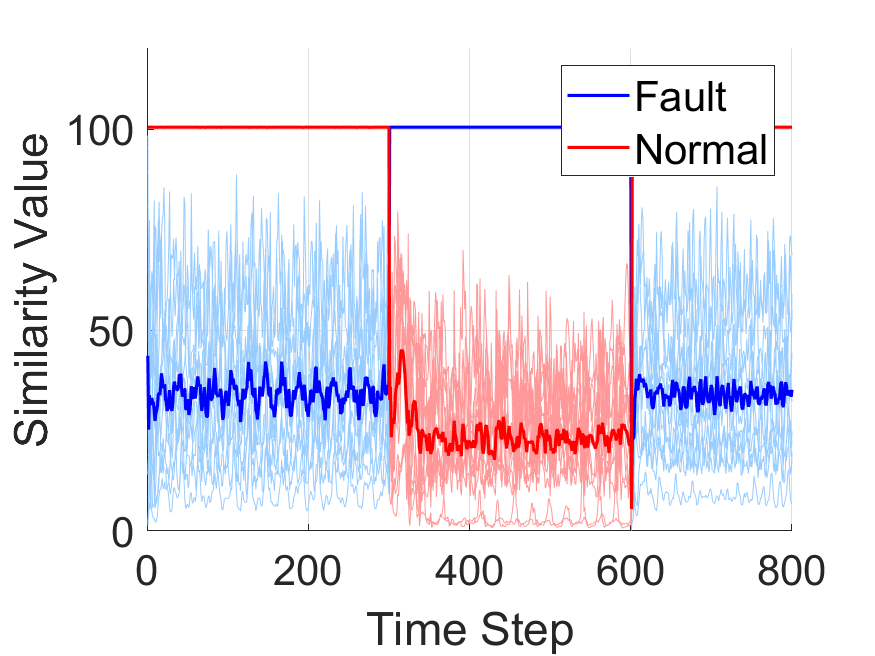}
    }
    \subfigure[Faults on Right Front Leg]{
        \includegraphics[width=0.45\linewidth]{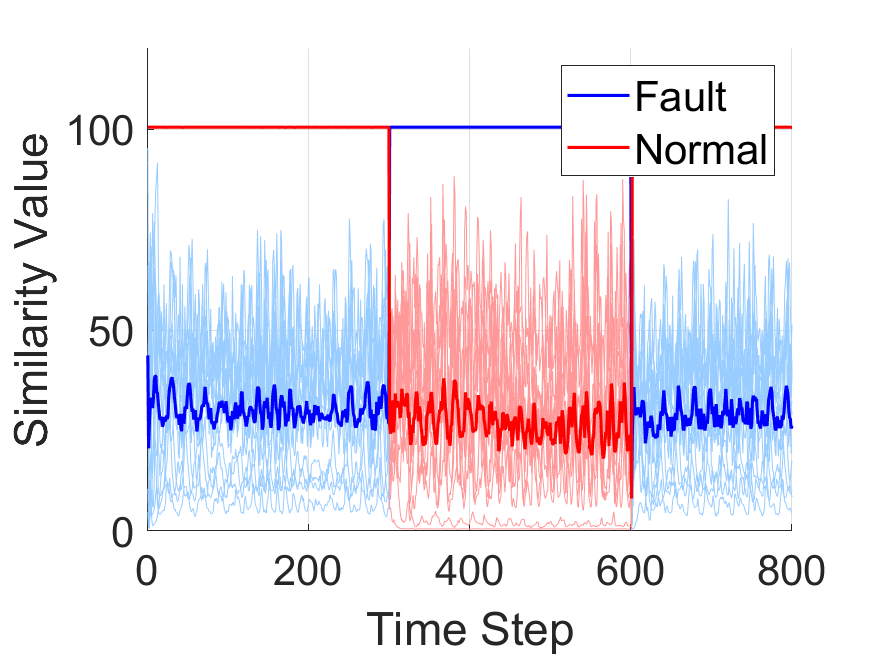}
    }
    \subfigure[Faults on Left Rear Leg]{
        \includegraphics[width=0.45\linewidth]{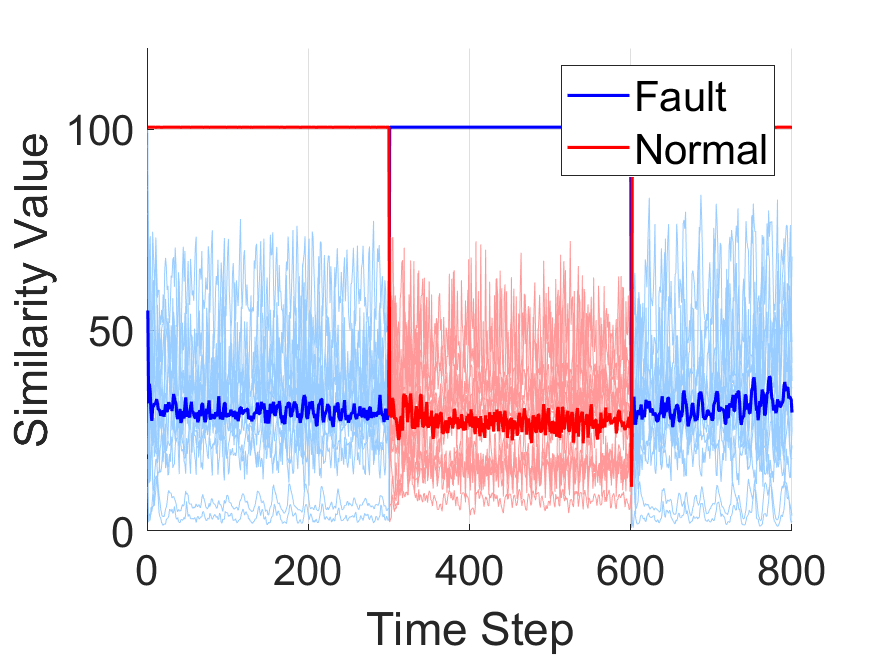}
    }
    \subfigure[Faults on Right Rear Leg]{
        \includegraphics[width=0.45\linewidth]{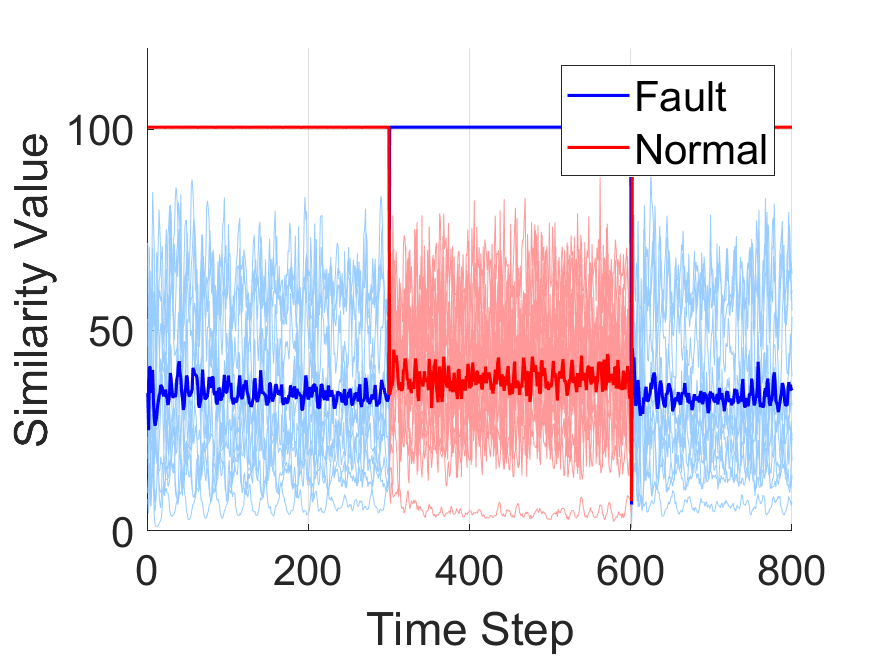}
    }
    \caption{Style reward evolution during locomotion tasks under fault conditions, indicating the switching of gait patterns. Blue curves: style rewards evaluated between the learned policy and the teacher police of the corresponding faulty case; Red curves: style rewards evaluated between the learned policy and the teacher policy of the normal case. Faults are introduced at step 300 and removed at step 600. For each leg, with one or two faulty joints, there are six possible fault scenarios. Reward values are generated for all scenarios, and the mean values are represented by bold curves.}
    \label{fig:combinedfaultsstyles}
\end{figure}

\subsection{Input to the Policy Network}
The observation space we used both in the simulation and the real cases only consist of the proprioceptive observations of the quadruped, which is totally collected by the onboard sensors on the quadruped. Thus, state estimations and cameras are not needed for the deployment.
\begin{equation}
o_{\text{proprioceptive}} = 
\begin{bmatrix}
\omega_{\text{base}}^\top & g^\top & p_{\text{DoF}}^\top & \omega_{\text{DoF}}^\top 
\end{bmatrix}^\top ,
\end{equation}
where $\omega_{\text{base}}\in\mathbb{R}^3$ is the angular velocity of the baselink, $g\in\mathbb{R}^3$ is the gravity, $p_{\text{DoF}}\in\mathbb{R}^{12}$ is the position of each joint and $\omega_{\text{DoF}}\in\mathbb{R}^{12}$ is the velocity of each joint.

Commands and actions are also given as input. Concatenate all the components together, we get the following input for the policy network.
\begin{equation}
\text{Input} = 
\begin{bmatrix}
o_{\text{proprioceptive}}^\top & c^\top & a_{\text{previous}}^\top 
\end{bmatrix}^\top , \label{eq:input}
\end{equation}
where $c\in\mathbb{R}^3$ is the velocity command given to the quadruped and $a_{\text{previous}}\in\mathbb{R}^{12}$ is the action computed by the policy at the last step.

We additionally adopted a mini-batch for training, together with the time-sequence data for the RNN, the final input to the student policy network has a shape of $(N, S, D)$, where $N$ is the batch size (in this task, equals to the number of the parallel environments), $S$ is the length of the time input sequence, and $D$ is the feature dimension (in this task, equals to the dimension of the Input \eqref{eq:input}).

\subsection{Reward Design}
To train multiple teacher policies for handling various fault cases, we design a set of reward functions tailored to each case to promote smooth and feasible gaits (see~\Cref{subsec:Reward} and~\Cref{reward}). Each teacher policy is optimized using case-specific reward functions with different types and numerical scales, reflecting the distinct motion requirements of each scenario.

To guide the decoder training of the student policy, we introduce two style rewards that help the student policy learn multiple gaits across different fault cases under the supervision of teacher policies, based on the similarity between the student policy action and the teacher policy action. The first style reward encourages the student policy to mimic the motion of the teacher policy by minimizing the Euclidean distance between their actions:
\begin{equation}
r_{\text{scale}} = e^{- \|a - a^*\|},
\end{equation}
where $a$ is the true action taken by the student policy and $a^{*}$ is the reference action generated by the corresponding teacher policy. Since the student’s actions are unlikely to exactly match the teacher’s policy, this reward helps promote the alignment in the action amplitude—specifically, the position command amplitude, which also affects the torque command amplitude—between the student and teacher policies.

Moreover, since we aim for the quadruped robot to learn a variety of limping gaits that require a broad range of torque outputs, directly copying the teacher policies will be inefficient. This is because the reward for more challenging gaits (e.g., walking with two faulty legs, which requires large position command with higher torque level output) tends to be small at the early stage of training (they are difficult to learn). To address this, we introduce a second style reward based on the cosine similarity between actions, which additionally encourages alignment in action direction.
\begin{equation}
r_{\text{direction}} = e^{\frac{a \cdot a^*}{\|a\| \|a^*\|}}
\end{equation}
This reward accelerates the initial learning process, helping the quadruped acquire workable yet suboptimal motions before gradually refining them into the desired gaits.

The reference actions for both style rewards are dynamically generated by the corresponding teacher policy. When the fault case changes, the appropriate teacher policy is selected to provide new reference actions.

To ensure robust performance across various terrains and gait transitions, the student policy must address several challenges not included in the teacher policies. Notably, teacher policies are trained without considering smooth transitions between different gaits, as they are initialized from static poses with fixed fault joints. Moreover, teacher policies are typically trained on specific flat terrains, whereas the deployed quadruped is expected to navigate more diverse environments.

Therefore, several regularization rewards are incorporated into the training to encourage the policy to perform stable motions. These regularization rewards play a critical role in consolidating all gaits into a single unified policy, enhancing robustness, and enabling adaptive fault-tolerant locomotion across multiple terrains.

Overall, the total reward is given by $r$:
\begin{equation}
r = c_{1}r_{\text{scale}} + c_{2}r_{\text{direction}} + c_{3}r_{\text{regularization}},
\end{equation}
where $r_{\text{regularization}}$ denotes the aforementioned regularization reward, and $c_{i},\ i=1,2,3$ are weighting hyperparameters.

\section{Experiments}
\subsection{Environment and Training Details} \label{subsec:Reward}
We adopt Isaac Gym\cite{makoviychuk2021isaac} as the simulation platform for the RL training, which enables GPU-based physics simulation and supports massively parallel training. The RL algorithm employed is Proximal Policy Optimization (PPO)\cite{schulman2017proximal}. The learned policies are tested and deployed on a Unitree Go2 quadruped robot.

Teacher policies are trained using an Actor-Critic paradigm with PPO, where both the Actor network and the Critic network are implemented as multi-layer perceptrons (MLPs).

\begin{figure}[t]
    \centering
    \includegraphics[width=0.98\linewidth]{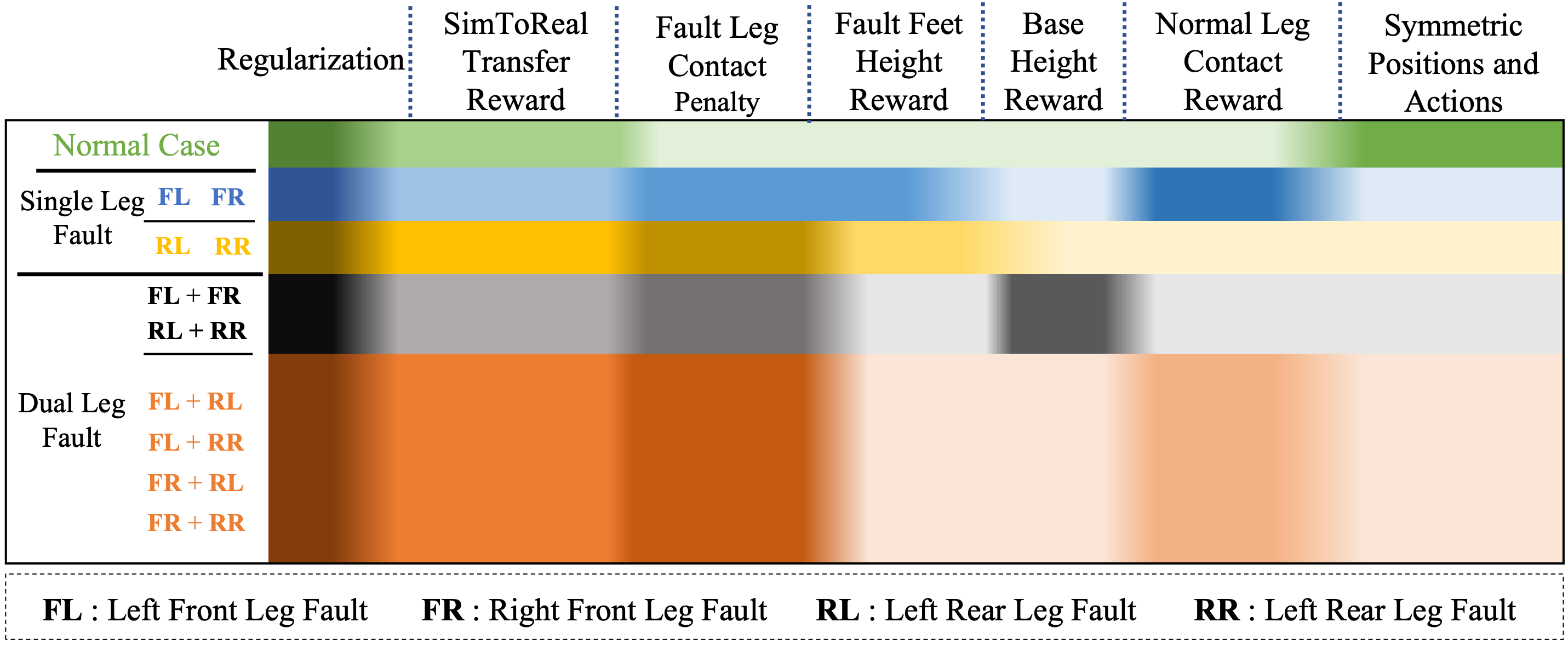}
    \caption{Reward design for teacher policy training. The color intensity represents the relative significance of each reward in shaping the gait — darker shades indicate higher importance and larger weights.}
    \label{reward}
\end{figure}

To tailor the teacher policies for different fault scenarios, we design custom reward functions for each case. These rewards include:
\begin{itemize}
    \item Penalizing ground contact by the faulty leg
    \item Encouraging the faulty leg to lift off
    \item Rewarding the base link height to maintain a stable posture
    \item Promoting ground contact by the healthy legs
\end{itemize}
The reward types and their relative significance are summarized in~\Cref{reward}. The numerical scales of these rewards are fine-tuned independently for each fault case to ensure the resulting gaits are both feasible and stable.

\begin{figure*}[!htbp]
\centerline{\includegraphics[width=0.66\linewidth]{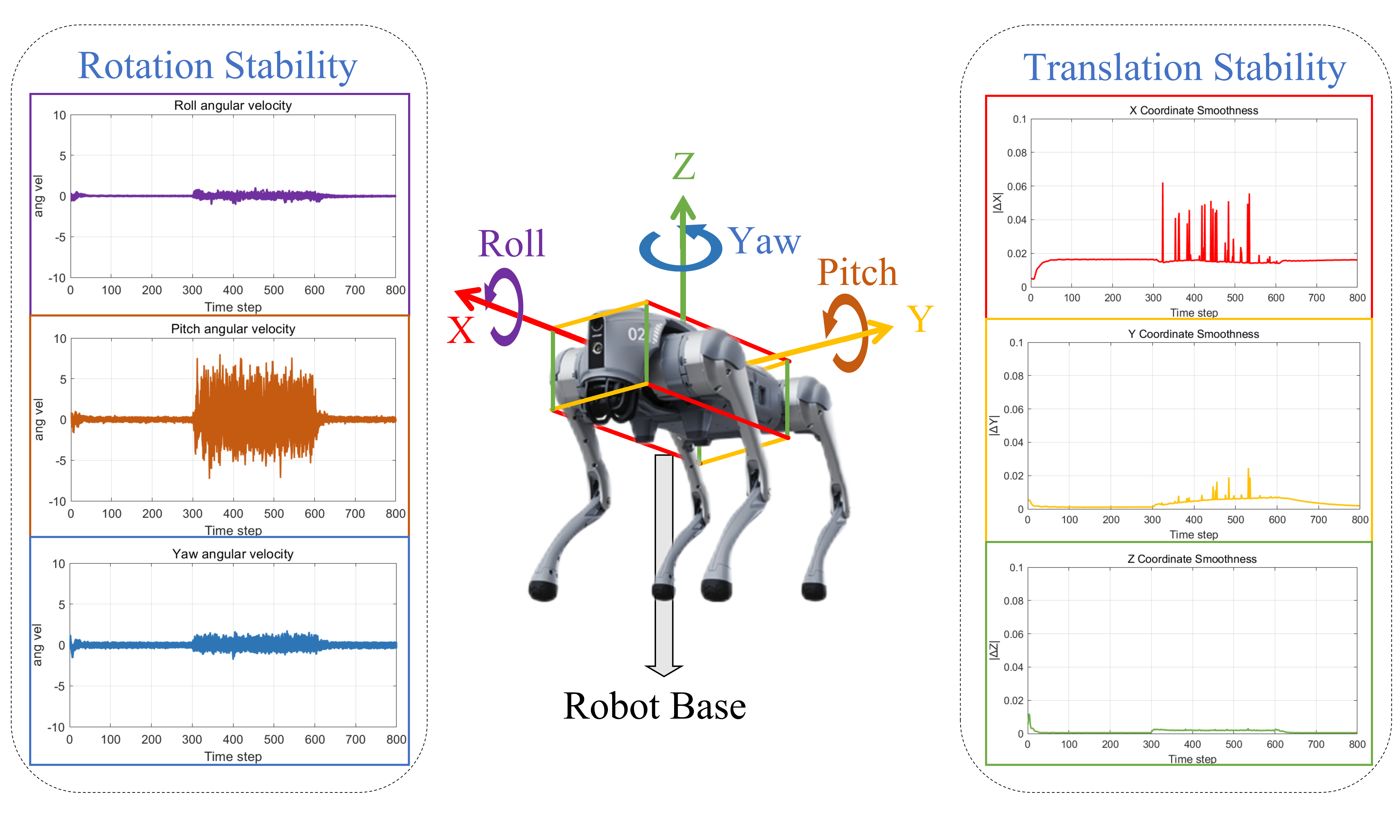}}
\caption{Rotation stability and translation stability of quadruped during a locomotion task with joint faults. Faults are introduced at step 300 and released at step 600. The stable evolution of the angular velocities and linear velocities indicate overall system stability.}
\label{figbaseheightplot}
\end{figure*}

Domain randomization techniques are employed during training to enhance the robustness of the policies. In simulation, parameters such as the terrain friction coefficient, base link mass, and motor output disturbances are randomized. Additionally, intermittent push forces are applied to the quadruped to improve its disturbance rejection capability. Detailed descriptions of the randomization types and their ranges are provided in the Appendix.

After training in Isaac Gym, the teacher policies are transferred to Gazebo~\cite{koenig2004design} to evaluate their gait transition performance in a different simulation environment. Thanks to independent fine-tuning, the teacher policies can be directly deployed on the physical quadruped and achieve reliable performance without further adaptation.

The student policy is trained using the Actor-Critic PPO algorithm, with hyperparameters identical to those used for the teacher policies. Domain randomization methods are consistently applied during student policy training to improve robustness across various conditions.

To facilitate fault-tolerant locomotion and smooth gait transitions, fault cases are switched multiple times within each training epoch. Specifically, fault joints are randomly changed every 300 simulation steps, and the corresponding teacher policy is simultaneously switched to provide style rewards. This approach enables the quadruped to learn how to maintain balance under sudden faults and transition seamlessly between different gaits. The time step is not included in the policy input, ensuring that the case-switching interval does not affect the learned policy.

To simulate torque-losing faults, the motor output torque of the affected joints is manually set to zero in both simulation and real-world experiments.

\subsection{Results}
We first evaluate the performance of the teacher policies in both Gazebo and real-world scenarios. The results from the Gazebo simulation are shown in Figure~\ref{fig:combinedteacherpolicies}. The training for each fault case separately results in excellent performance, with the quadruped exhibiting natural gaits under all conditions.

The trained policy is then deployed on a real robot, demonstrating robust performance in several key aspects:

\begin{itemize}
    \item The quadruped can execute normal gait as well as fault-tolerant gaits. 
    \item It can automatically transition smoothly between gaits when a fault occurs.
    \item  The quadruped is capable of completing locomotion tasks on uneven terrain. 
    \item It can return to its normal gait once recovery from a fault is achieved. 
\end{itemize}

\subsubsection{Switching between gaits without additional commands}
We simulate a faulty case by manually setting some torques to zero. The quadruped adapts rapidly to the new fault condition and transitions between gaits automatically, without the need for upper-level commands.

Figure~\ref{fig:combinedfaultsstyles} illustrates the variation in style rewards for different teacher policies corresponding to the normal and faulty cases, respectively. During locomotion tasks with faults, the style reward between the policy and the teacher policy for the normal gait drops sharply while that for the corresponding limping gait rises quickly. Conversely, when the fault is disabled, the style reward for the normal gait rapidly increases and that for the limping gait rapidly decreases. This indicates that gait transitions occur promptly in response to fault conditions.

Moreover, Figure~\ref{figbaseheightplot} illustrates the quadruped's stability during a locomotion task in which faults are introduced and subsequently removed, showcasing its ability to maintain stability throughout the bidirectional gait transitions.

\subsubsection{Resistance, robustness and stability}
To further assess the robustness of the fault-tolerant locomotion policy, we introduce external disturbances to the quadruped. The robot exhibits the ability to resist kicks and pushes, recovering quickly from the disturbances (See \Cref{figtitlepic}(b)). Furthermore, the quadruped can walk and smoothly transition between gaits while carrying unsecured parcels on its back without dropping them (See \Cref{figtitlepic}(c)).

\subsection{Ablation Studies}

\begin{figure}[!htbp]
    \centering
    \includegraphics[width=0.95\linewidth]{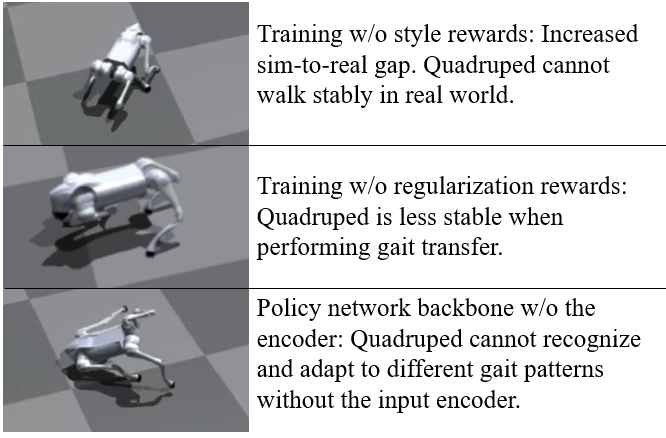}
    \caption{Ablation studies of our proposed learning framework}
    \label{fig:combinedbaselines}
\end{figure}

We conducted three ablation studies to assess the effectiveness of our design. To ensure fair comparisons, all other training settings were kept identical across the three cases. The results are presented in~\Cref{fig:combinedbaselines}.

\subsubsection{Training w/o style rewards} 
The policy is trained without the style rewards using the same checkpoint as the AcL. The rewards used in the training of teacher policies are added to replace the style rewards in this section. Without the style rewards, the quadruped performs gaits that are not similar to the teacher policies, and not feasible to transfer to real cases in the simulation.

\subsubsection{Training w/o regularization rewards}
The student policy is trained only with the style rewards, and the regularization rewards are removed. Without the regularization rewards, the quadruped is not able to learn locomotion tasks in training when fault joints are enabled in the simulation.

\subsubsection{Policy network backbone w/o the encoder}
The encoder in the policy network structure is removed. Without the encoder in the policy network backbone, the quadruped cannot identify different fault cases, which leads to a failure when faults are enabled.

\section{Conclusions}

We proposed Action Learner (AcL) — a novel teacher-student reinforcement learning framework that enables a quadruped to achieve adaptive fault-tolerant locomotion under multiple joint faults. The trained policy allows the quadruped to maintain stable locomotion and smoothly transition between gaits, even when faults occur on multiple joints across one or two legs. With the design of the style reward and encoder-decoder architecture, the framework distills various fault-specific gaits into a single policy, eliminating the need to train separate policies for different fault scenarios.

The modular training framework facilitates independent fine-tuning of individual teacher policies, resulting in more feasible and efficient motions for the quadruped. The resulting student policy achieves optimal performance across diverse fault cases and environments. Additionally, the framework is readily extendable to incorporate new tasks via recovery training. For example, teacher policies designed for stair-climbing, slope-climbing, or step-stone crossing can be integrated, allowing the quadruped to recognize terrains and adaptively adjust its gait.

Moreover, the framework is agnostic to the choice of teacher policies, provided they operate under position control. This flexibility enables the seamless integration of model-based controllers into the training pipeline, allowing the student policy to inherit more symmetric and elegant motions while maintaining robust fault-tolerant capabilities.

\appendix
\section{Training Details} \label{sec:app}

~\Cref{tab:networkstructure} below shows the structure of the student actor networks and the corresponding training hyperparamters.
\begin{table}[htbp]
\caption{Student Actor Network and Training Settings}
\begin{center}
\begin{tabular}{l|l}
\hline
Input time sequence   length $S$ & 10                                       \\
Encoder dims                & {[}512,512,512{]}                        \\
Encoder input dim         & {[}1,10,45{]}                              \\
Encoder output dim         & {[}4{]}                              \\
Decoder dims                & {[}512,256,128{]}                        \\
Decoder input dim              & {[}1,49{]}                        \\
Critic network dims         & {[}512,256,128{]}                        \\
Number of environments $N$      & 3456                                     \\
Training iterations         & $>$5000                                  \\
Reward scale $c_{1}$ & 100 \\
Reward scale $c_{2}$ & 5 \\ \hline
\end{tabular}
\end{center}
\label{tab:networkstructure}
\end{table}

~\Cref{tab:domainrand} below shows the domain randomization setting used in the training of the student policy.
\begin{table}[htbp]
\caption{Domain Randomization}
\begin{center}
\begin{tabular}{l|l}
\hline
Randomize friction   range   & {[}0.5,1.25{]}                        \\
Randomize base mass range    &  {[}-1.0,1.0{]}                       \\
Push forces velocity         &  1.0                                  \\
Push forces interval         & 15                                    \\
False switch steps           & 300                                   \\
Randomize motor output range & {[}0.8,1.2{]}                         \\ \hline
\end{tabular}
\end{center}
\label{tab:domainrand}
\end{table}

\bibliographystyle{plain}
\bibliography{sample}

\end{document}